\documentclass[runningheads]{llncs}

\usepackage[long]{optional}
\usepackage{todonotes} 

\usepackage[utf8x]{inputenc}
\usepackage[american]{babel}
\usepackage{csquotes}

\usepackage{graphicx}
\usepackage{mathptmx}
\usepackage{amsmath,amssymb,amsfonts}

\usepackage[hidelinks,pdftex,unicode]{hyperref}
\usepackage{url}
\usepackage[capitalize,noabbrev]{cleveref}
  \creflabelformat{equation}{#2#1#3}

\usepackage{subcaption}
\usepackage{multirow}
\usepackage{wrapfig}

\begin{document}

\title{Fast Classification Learning with Neural Networks and Conceptors
	for Speech Recognition and Car Driving Maneuvers}
\titlerunning{Fast Classification Learning with Recurrent Neural Networks}

\author{Stefanie~Krause\inst{1,2}\orcidID{0000-0002-1271-7514} \and
Oliver~Otto\inst{2} \orcidID{0000-0002-8982-7164} \and
Frieder~Stolzenburg\inst{2}\orcidID{0000-0002-4037-2445}}
\authorrunning{S.~Krause \and O.~Otto \and F.~Stolzenburg}
\institute{
	University of Bayreuth,
	Research Center Finance \& Information Management,
	\opt{long}{Wittelsbacherring 10,
	95444 }Bayreuth, Germany,
	\email{stefanie.krause@fim-rc.de},
	\url{http://www.fim-rc.de/en/}
\and
	Harz University of Applied Sciences,
	Automation and Computer Sciences Department,
	\opt{long}{Friedrichstr.~57--59,
	38855 }Wernigerode, Germany,
	\email{\{ootto,fstolzenburg\}@hs-harz.de},
	\url{http://artint.hs-harz.de/}}

\maketitle

\begin{abstract}
Recurrent neural networks are a powerful means in diverse applications. We show
that, together with so-called conceptors, they also allow fast learning, in
contrast to other deep learning methods. In addition, a relatively small
number of examples suffices to train neural networks with high accuracy. We
demonstrate this with two applications, namely speech recognition and detecting
car driving maneuvers. We improve the state of the art by application-specific
preparation techniques: For speech recognition, we use mel frequency cepstral
coefficients leading to a compact representation of the frequency spectra, and
detecting car driving maneuvers can be done without the commonly used polynomial
interpolation, as our evaluation suggests.
\end{abstract}

\section{Introduction}
The field of artificial intelligence nowadays is dominated by machine learning
and big data analysis, in particular with deep neural networks\opt{long}{ \cite{GBC16}}.
Deep learning in general means a class of machine learning algorithms that use a
cascade of multiple layers of nonlinear processing units for feature extraction
and transformation \cite{DY14}. The tremendous success of deep learning in
diverse fields of artificial intelligence such as computer vision and natural
language processing seems to depend on a bunch of ingredients: deep networks
with nonlinearly activated neurons, convolutional layers, and iterative training
methods like backpropagation.

Since deep neural networks often consist of thousands of neurons, the
corresponding learning procedures require a large number of training examples,
usually several thousands per example class for classification tasks (cf., e.g.,
\cite{War18}). Otherwise the networks do not generalize well, and overfitting
is a problem. So it would be nice to work with smaller networks, which also
reduces the time for training neural networks significantly. For this, we
investigate small recurrent neural networks (RNNs) and their applicability to
different tasks.

We demonstrate that a relatively small number of examples suffices to train
neural networks with high accuracy by two applications, namely speech
recognition and detecting car driving maneuvers. We employ RNNs, more precisely echo state networks (ESNs) with conceptors
\cite{Jae07,Jae14,Jae17}. Training them is very fast, and they can effectively be
learned by simply solving a linear equation system, backpropagation or similar
methods are not needed. Moreover, we improve the state of the art by
application-specific preparation techniques.

\section{Background and Related Works}

\subsection{Recurrent Neural Networks}\label{sec:rnn}

\opt{long}{RNNs can be used to create temporal dependencies, which occur in many
sequence modeling tasks \cite{SL+18}. Nevertheless, RNNs are not the only models
capable of representing time dependencies. Hidden Markov models (HMMs), that
develop an observed sequence as probabilistically dependent upon a sequence of
unobserved states, can be used as well \cite{al2018deep}. However, RNNs overcome
the main limitation of Markov models by capturing long-range time dependencies
\cite{al2018deep}. They contain recurrent connections that make them a powerful
means to model sequential data \cite{sak2014long}. RNNs can maintain an
activation even in the absence of input and thus exhibit dynamic memory
\cite{jaeger2004harnessing}. A general RNN in contrast to a feed-forward network
is shown in \cref{bild1}. The key advantages of RNNs can be seen quickly:
Feedback from the output back to the hidden states is possible, as well as
recurrent connections in the hidden states itself.

\begin{figure}
	\centering
	\includegraphics[
	clip=true,
	trim= 0 0 14cm 0,height=5cm
	]{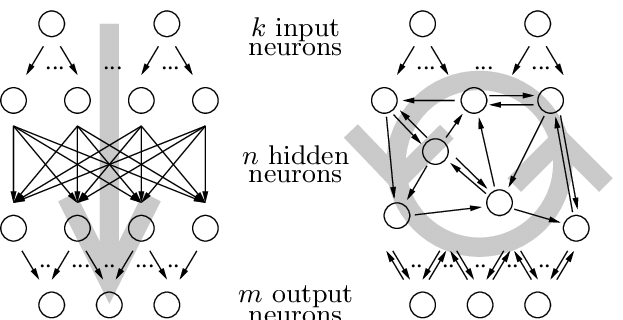}%
	\hspace{2em}
	\includegraphics[clip=true,
	trim= 12.5cm 0 0 0, height=5cm
	]{pics/RNNandfeedword.png}
	\caption{Comparison of the architecture of a feed-forward network (left) with a
	recurrent neural network (right). The gray arrows show the direction of
	computation. Feed-forward networks compute only in one direction in
	contrast to RNNs which have recurrent dependencies (cf.
	\cite{burgsteiner2007movement}).}
	\label{bild1}
\end{figure}}

We now briefly introduce RNNs, following the lines of \cite{Jae14}. We
use discrete-time RNNs where time is progressed in unit steps $n = 1,2,\dots$
and $\tanh$ (hyperbolic tangent) as activation function which squashes the neuronal
activation values into a range between $-1$ and $+1$. For a network consisting of $N$
neurons, the activations $x_1(n),\dots, x_N(n)$ at time $n$ are collected in an
$N$-dimensional state vector $x(n)$. The \opt{long}{(at least in the beginning)
}random weights of the neuron\opt{long}{ connection}s are
collected in a weight matrix $W^{\text{res}}$ of size $N \times N$. $p(n)$ is the input
signal fed to the network, where the input weights are collected in the weight
matrix $W^{\text{in}}$ of size $N \times d$ (for a $d$-dimensional input). $b$
is a bias. The network update equation is:
\begin{equation}
x(n+1)= \tanh\left(W^{\text{res}} x(n) + W^{\text{in}} p(n+1)+b\right)
\label{eq:update}
\end{equation}
The network-internal neuron-to-neuron connections comprised in the matrix
$W^{\text{res}}$ are (and remain) random. This includes the existence of cyclic
(recurrent) connections. The equation
\begin{equation}
y(n)=W^{\text{out}} x(n)
\label{eq:out}
\end{equation}
specifies that an output signal $y(n)$ can be read from the network activation
state $x(n)$ via the output weights comprised in the matrix $W^{\text{out}}$ of
size $d \times N$. These weights $W^{\text{out}}$ are computed by simply solving
the linear equation system of \cref{eq:out} such that the output signal $y(n)$
is the prediction of the (next) input signal $p(n)$, i.e., $y(n)=p(n+1)$.

\subsection{Echo State Networks and Conceptors}

ESNs provide a supervised learning principle for RNNs.\opt{long}{ The basic
ESN network architecture is shown in \cref{esn}.}
They employ a randomly connected neural network with fixed random weights,
called reservoir.\opt{long}{ In \cref{esn} the reservoir coincides with
the $N$ internal units with the random fixed weights in $W^{\text{res}}$.} Only the
connections from the reservoir to the output readout neurons are modified in the
training phase by learning \cite{jaeger2004harnessing}.\opt{long}{ This is
possible because of the echo state property. It says that any random initial
state of a reservoir is \enquote{forgotten} after a washout period such that the
current network state is a function of the driver \cite{Jae14}.} The input
signal of an ESN induces a non-linear response signal in each neuron within the
reservoir network. A linear combination of these response signals is trained to
obtain the desired output \cite{Jae14,jaeger2004harnessing}. That is why
training an ESN becomes a simple linear regression task
\cite{jaeger2004harnessing} (cf. \cref{sec:rnn}).

\opt{long}{\begin{figure}
	\centering
	\includegraphics[height=4.8cm]{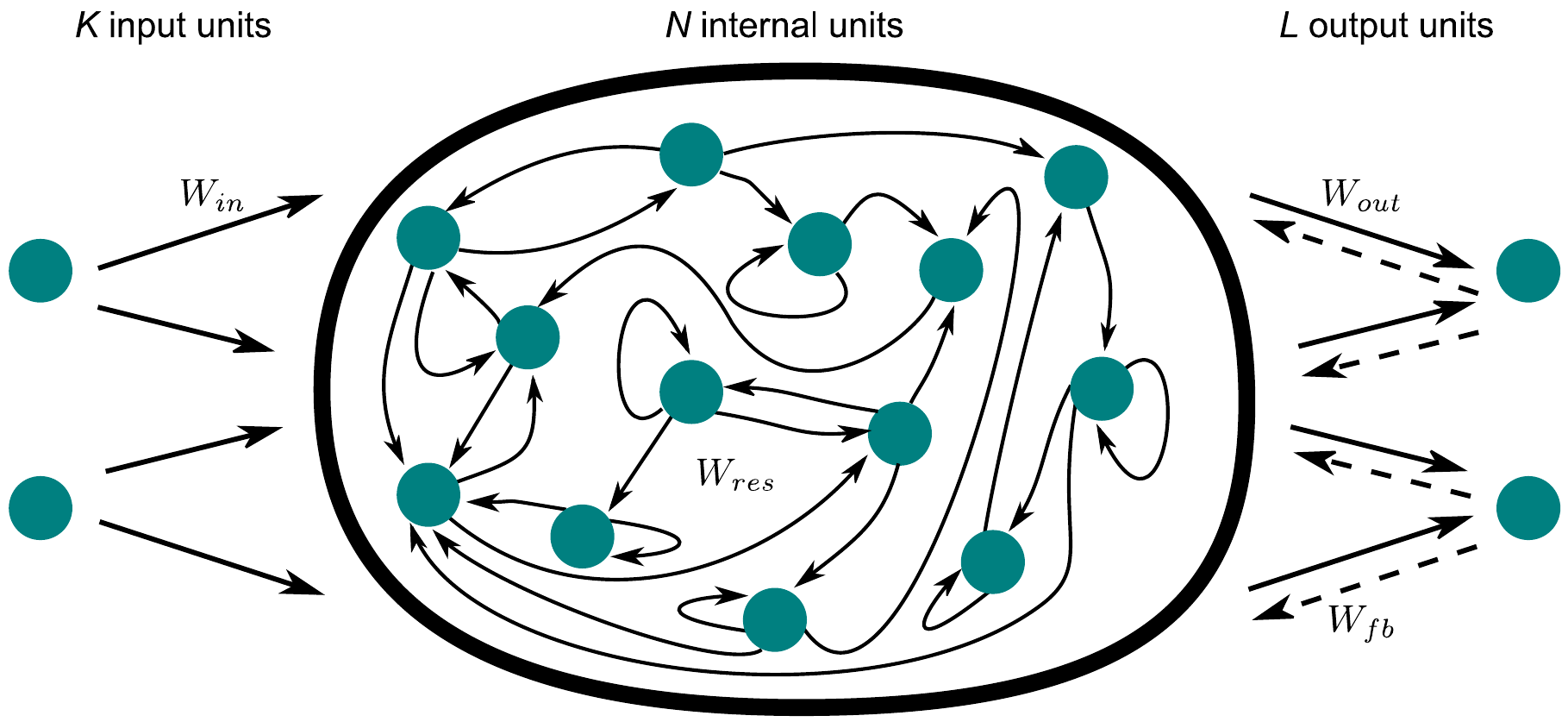}
	\caption{The basic network architecture of an echo state network (cf. \cite{jaeger2001echo}). Dashed
		arrows indicate connections that are possible but not required.}
	\label{esn}
\end{figure}

According to \cite{jaeger2002tutorial}, ESNs can be generated by the following procedure:
\begin{enumerate}
	\item Generate a random internal weight matrix $W_0$.
	\item Normalize $W_0$ to a matrix $W_1$ with unit spectral radius by putting
	$W_1 = 1/ \vert \lambda_{\text{max}} \vert \cdot W_0$,
	where $\lambda_{\text{max}}$ is the spectral radius of $W_0$.
	\item Scale $W_1$ to $W^{\text{res}} = \alpha \cdot W_1$, with $\alpha < 1$, whereby $W^{\text{res}}$ has a spectral radius of $\alpha$.
	\item The untrained network ($W^{\text{in}}, W^{\text{res}}, W^{\text{out}}$) is then an ESN, regardless of how $W^{\text{in}}$, $W^{\text{out}}$ are chosen.
\end{enumerate}}

Conceptors, their basic mechanisms and characteristics and a mathematical
definition are described in \cite{\opt{long}{jaeger2014conceptors,}Jae14,Jae17}.
In \cite[Sect.~3.13]{Jae14}, the classification of Japanese vowels by means of
conceptors is considered: Nine male native speakers pronouncing the Japanese
di-vowel /ae/ should be recognized. The corresponding procedure employing
conceptors has served as the basis for the tasks considered here: speech
recognition and car driving maneuvers classification.
\opt{long}{\cite{Jae17} focuses on neural long-term memory for
temporal patterns. In that paper, conceptors are described more general as a
neuro-computational mechanism which can be used in diverse neural
information processing tasks like neural noise suppression, stabilization of
neural state dynamics and signal separation.}

Let us now introduce conceptors in some more detail, following the lines of \cite{Jae14}:
A conceptor matrix $C$ for some vector-valued random variable $x \in
\mathbb{R}^N$ is defined as a linear transformation that minimizes a loss function:
\begin{equation}C(R, \alpha) = \textnormal{argmin}_C\; E \left[ \left\| x - Cx\right\| ^2 \right] + \alpha^{-2} \left\|  C\right\| ^2 _{\text{fro}}
\label{eq:eins}
\end{equation}
where $\alpha $ is a control parameter, called aperture, $E$ stands for the
expectation value (temporal average), and $|| \cdot ||_{\text{fro}}$ is the
Frobenius norm. The closed-form solution for this optimization problem is
\begin{equation}
C(R, \alpha) = R(R + \alpha^{-2} I)^{-1} = (R+ \alpha^{-2} I^{-1})R
\label{eq:zwei}
\end{equation}
where $R=E \lbrack xx^{'} \rbrack$ is the $N \times N$ correlation matrix of
$x$, and $I$ is the $N \times N$ identity matrix. \opt{long}{Note that the inverses
appearing in \cref{eq:zwei} are well-defined because $\alpha > 0$ is
assumed, which implies that all singular values of $C(R, \alpha)$ are properly
smaller than~$1$.}

To understand \cref{eq:zwei}, we have to look at the {singular value
decomposition} (SVD) of $C$. If $R = U \Sigma V$ is the SVD of $R$ (where
$\Sigma$ is a rectangular diagonal matrix with non-negative real numbers on the
diagonal), then the SVD of $C(R,\alpha)$ can be written as $USV$ where the
singular values $s_i$ of $C$ (comprised in the diagonal matrix $S$) can be written in terms of the
singular values $\sigma_i$ of $R$: $s_i = \sigma_i /(\sigma_i + \alpha^{-2}) \in
[0, 1)$, for $\alpha \in (0, \infty)$.

\cite{Jae14} finds that the non-linearity inherent in \cref{eq:eins}
makes the conceptor matrices come out almost as projector matrices because the
singular values of $C$ are mostly close to $1$ or $0$. In intuitive terms, $C$
is a soft projection matrix on a linear subspace, where samples $x$ are
projected into this subspace. $C$ almost acts like the identity: $Cx \approx x$.
\opt{long}{When some noise $\xi$ orthogonal to the subspace is added to $x$, $C$
reconstructs $x$ by filtering the noise: $C(x + \xi) \approx x$.}

Conceptor matrices are positive semi-definite matrices whose singular vectors
are the axes of the conceptor ellipsoids. Their singular values in the unit
interval $[0,1]$ represent the lengths of the ellipsoid axes.
\opt{long}{Three examples of different patterns with their corresponding conceptor ellipsoids are shown in \cref{conceptors}.}
In practice, the correlation matrix $R$ is estimated from a finite sample $X =
(x(1), \dots, x(L))$ where the $x(n)$ are the reservoir states collected during a learning
run. This leads to the approximation $R_{\text{approx}} = XX^T/L$. An optimal
aperture $\alpha$, which can be interpreted as a scaling factor, can be found by
a cross-validation search (cf.~\cite{he2018overcoming}). Using the correlation
matrix $R_j$ from a pattern $p_j$ and $\alpha$, one can compute the conceptor
matrix $C^j = C(R_j, \alpha)$.

\opt{long}{
\begin{figure}
	\centering
	\includegraphics[height=3.5cm]{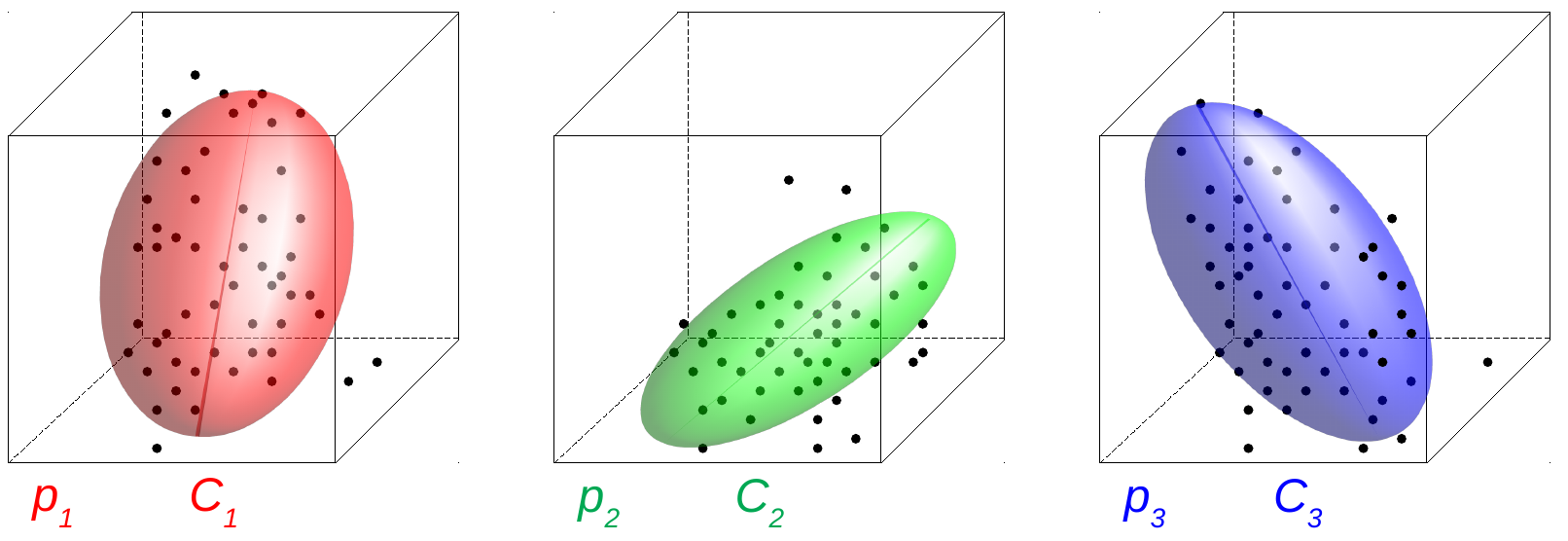}
	\caption{Conceptors $C^1, C^2, C^3$  of three pattern $p^1, p^2, p^3$ of a network with $N=3$ reservoir neurons.}
	\label{conceptors}
\end{figure}}

\section{Classification with Conceptors}

\subsection{Conceptor Algebra}
On matrix conceptors, operations that satisfy most laws of Boolean logic such
as NOT ($\lnot$), OR ($\lor$), and AND ($\land$) can be defined
\cite{he2018overcoming}\opt{long}{, as shown in \cref{bild2}}. Although not all
laws of Boolean algebra are satisfied, these operations bring a great advantage
for conceptors, because if we get new patterns we need not train everything
from the start, but use overlaps with other already saved patterns and just save
the new components (which saves memory space)~\cite{Jae14}. Furthermore, we can
use Boolean operations to check how many patterns lie in one conceptor and not
in any other and for similar considerations.

\opt{long}{
	\begin{figure}
		\centering
		\includegraphics[width=8.8cm,height=3.3cm]{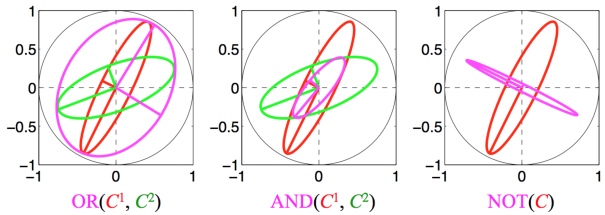}
		\caption{Boolean operations on 2-dimensional conceptors are shown (cf. \cite{he2018overcoming}). Red/green ellipses represent source conceptors $C^1$, $C^2$ and the pink ellipses show $C^1 \lor C^2$, $C^1 \land C^2$, $\lnot C^1$ from left to right.}
		\label{bild2}
\end{figure}}

\subsection{Classification}\label{evidence}

To use conceptors for a classification task, we first determine two conceptors,
called positive and negative evidence, for each class, e.g., words, speakers, or
car driving maneuvers, based on the training data. For instance, for the
classification of nine speakers, we have twice as many conceptors. For every
class, we compute (train) one conceptor $C^j$ characteristic of class~$j$ and
determine a second conceptor
\begin{equation}\label{negevi}
N^j = \lnot (C^1 \lor \dots \lor C^{j-1} \lor C^{j+1}\lor\dots\lor C^m)
\end{equation}
where $\lnot$ is the Boolean negation (NOT), $\lor$ is the Boolean operation
disjunction (OR) and $m$ is the number of different classes. This characterizes
the condition that this class $j$ is not any of the other classes
$1,\dots,j-1,j+1,\dots,m$.

After the training phase, we can classify a sample by computing the evidence values
for each conceptor as well as a combined evidence. The class corresponding to the
conceptor with the highest evidence value is then assumed to be the correct
classification. We get the positive evidence of a conceptor by computing
$E^+(p,j)=x^TC^jx$ where $x$ is the response signal of the network. This leads
to a classification by deciding for $j= \text{argmax}_i\,x^T C^+_i x$ to select
class~$j$. $E^+(p,j)$ is a non-negative number indicating how well
the signal fits into the ellipsoid of $C^j$. The idea behind this is that, if
the reservoir is driven by a class~$j$, the resulting response signal $x$ will
be located in a linear subspace of the (transformed) reservoir state space whose
overlap with the ellipsoids $C_i$ $(i=1,\dots,m)$ is maximal for $i=j$.

Similarly, the negative evidence $E^-(p,j)=x^TN^jx$ (which has a positive value
despite its name) is computed. It describes that the sample $p$ is of class j
and not one of the other classes. Since $E^-(p,j)=x^TN^jx$ we can compute the negative
evidence of a pattern using the Boolean operations NOT and OR (cf. \cref{negevi}). To obtain the
combined evidence, we simply sum up the previous two values:
\begin{equation}\label{combined}
E(p,j)=E^+(p,i) + E^-(p,i)
\end{equation}
Thus the combined evidence has a greater value if both the positive
evidence $E^+(p,j)$ and the negative evidence $E^-(p,j)$ have high values. This
is the case if the signal is inside the conceptor $C^j$ and has low overlap with
any other conceptor.
For each utterance, we compute $m$ combined evidences
$E(p,1),\dots,E(p,m)$. The pattern $p$ can then be classified as class~$j$ by
choosing the class index $j$ whose combined evidence $E(p,j)$ was the greatest
among the collected evidences.

\section{Case Studies}

\subsection{Speech Recognition}

In this case study, we consider two different tasks: A.~recognition of isolated words and B.~speaker
recognition. We used speech samples of adults and children of age between $3$
and $45$ years, female and male. The datasets for both tasks overlap but are
not identical.

\begin{description}
\item[Isolated Word Recognition:]
For the recognition of isolated words we have a total of 1320 samples, divided
in a training set of 880 utterances (2/3 of all samples) and a test set of
440 utterances (1/3 of all samples). In this task we recognize the German isolated words
\enquote{halt}, \enquote{langsamer}, \enquote{links}, \enquote{rechts},
\enquote{schneller}, \enquote{stopp}, \enquote{vor}, and \enquote{zurück}
pronounced by different speakers (adults and children, female and male). For
each of our 8 different words we have 110 training and 55 test samples.

\item[Speaker Recognition:]
For this task, nine persons spoke different distinct isolated German words (per
speaker we used up to 26 different words). The number of speech samples differs
per speaker. The training set of every speaker is the same with 65 samples. The
amount of test samples varies per speaker. Overall we have 296 test
samples. We choose samples of children and adults, female and male speakers to
create a diverse dataset, with different speech rate and pronunciation.
\end{description}

The raw speech samples need to be preprocessed, before we can pass them to
the neural network. Speech recognition performance and sensitivity depends
heavily on preprocessing\opt{long}{ \cite{ibrahim2017preprocessing,shanthi2013review}},
because the amount of data supplied to the input of the neural network usually
is reduced by this. Preprocessing differentiates the voiced or unvoiced signal
and creates feature vectors \cite{ibrahim2017preprocessing}. We develop a
speaker and isolated word recognition system based on the Japanese vowels
classification system in \cite{Jae14} and employ the Matlab code from
\url{http://minds.jacobs-university.de/research/conceptorresearch/}.

We apply application-specific preparation techniques and divide the signal into
discrete sequences of feature vectors that only contain relevant information
about the utterance. For this, we extract features by employing mel frequency
cepstral coefficients (MFCCs) and obtain uncorrelated vectors by means of
discrete cosine transforms (DCT). MFCCs are the most used spectral feature
extraction method \cite{dey2019intelligent}. It is based on frequency analysis
using the mel scale similar to the human ear scale. The coefficients are a
representation of the real cepstral of a windowed short-time signal derived from
the fast Fourier transform (FFT) of that signal, more precisely, the logarithm
of the spectral amplitudes.\opt{long}{ The FFT algorithm extracts frequency
amplitudes for discrete intervals from a given segment of a signal
\cite{grewal2010isolated}. Examples of the MFCCs of two different German words
are shown in \cref{fig:MFCCs}.} In our experiments, we use 12 MFCCs and thus
obtain $12$-dimensional time series by means of a Python implementation using
the \texttt{librosa} package for music and audio analysis (see \url{http://librosa.org/}). Each utterance is sampled with
$512$ data points. Since the length of the original audio files is between $0.6$ and
$2.0$\,s, the sampling rate is hence between $256.0$ and $853.3$\,Hz.

\opt{long}{\begin{figure}
	\begin{subfigure}[b]{0.5\textwidth}
		\includegraphics[width=1\textwidth]{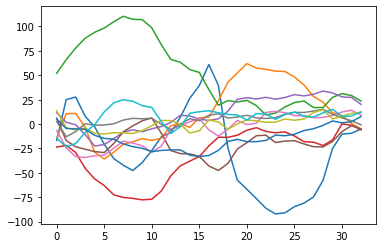}
		\caption{\enquote{halt}}
		\label{fig:halt}
	\end{subfigure}
	\begin{subfigure}[b]{0.5\textwidth}
		\includegraphics[width=1\textwidth]{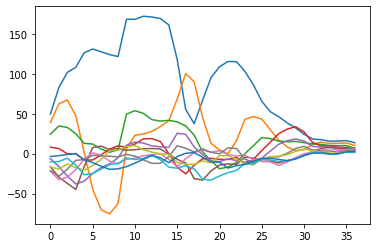}
		\caption{\enquote{langsamer}}
		\label{fig:langsamer}
	\end{subfigure}

	\caption{Selection of examples of the MFCCs from signals of the two
	German words \enquote{halt} and \enquote{langsamer}. From each utterance, we get a
	$12$-dimensional time series consisting of $12$ MFCCs. The $x$-axis of
	the plot corresponds to the sampling time steps (one frame consists of
	$512$ sampling points), and the $y$-axis to the MFCC values. As the
	exemplary utterances are from kids, which tend to have higher
	frequencies, the MFCC values are relatively high.}
	\label{fig:MFCCs}
\end{figure}}

\subsection{Car Driving Maneuvers}

In this case study, we investigate whether driving maneuvers can be classified
by the use of conceptors. Driving behavior has already been analyzed in
several papers, and possible applications have been presented
\cite{hong2014,IKL11,\opt{long}{johnson2011,SG+17,}TMW04,vaiana2014\opt{long}{,you2012}}. We
investigate here whether conceptors can also be used for car driving maneuver detection.
Furthermore, we consider the influence of the number of reservoir neurons and
whether very small training datasets are sufficient to train a conceptor.

The classification method we used for this task is based on the
dynamic pattern recognition method in \cite[p.\,74]{Jae14}. In addition, we have
investigated whether all components of the classification method in
\cite{Jae14} are actually necessary for the car driving maneuver classification task (cf. \cref{essen}).
The data was recorded with a smartphone which was
permanently installed in the car while driving. We took care of a fix
position of the smartphone relative to the vehicle. An app (written by the second author, cf. \cite{Ott20}) continuously
recorded the lateral, longitudinal, and gravitational acceleration acting on the
vehicle as well as the GPS data. The speed was derived from the GPS data. The
sampling rate was $10$\,Hz.

\begin{wraptable}[11]{r}{6.5cm}\vspace{-7mm}
  \caption{Overview of car driving maneuvers.}
  \label{tab:anzahlManoever}
  \small\centering
  \begin{tabular}{c|l|c}
	\hline
	\textbf{class index $j$} & \textbf{maneuvers}	& \textbf{number of samples} \\
	\hline
	1 & stop		& 10\\
	\hline
	2 & straight ahead		& 9\\
	\hline
	3 & start up			& 11\\
	\hline
	4 & slow down	& 8 \\
	\hline
	5 & full braking  	& 8 \\
	\hline
	6 & left turn		& 15 \\
	\hline
	7 & right turn		& 17\\
	\hline
  \end{tabular}
\end{wraptable}

\begin{figure}
	\begin{subfigure}[t]{0.39\textwidth}
		\includegraphics[width=1\textwidth]{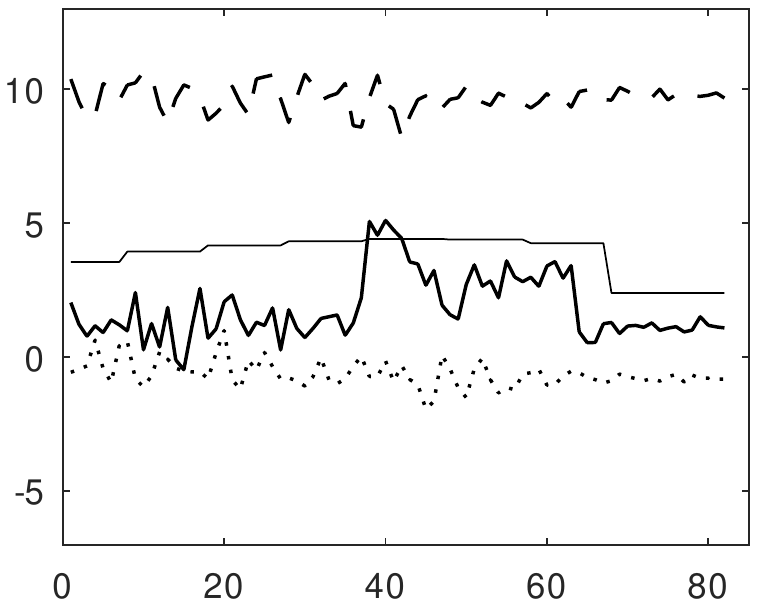}
		\caption{slow down}
		\label{fig:Messreihe_1_Normalbremsung}
	\end{subfigure}
	\opt{long}{
		\begin{subfigure}[t]{0.39\textwidth}
			\includegraphics[width=1\textwidth]{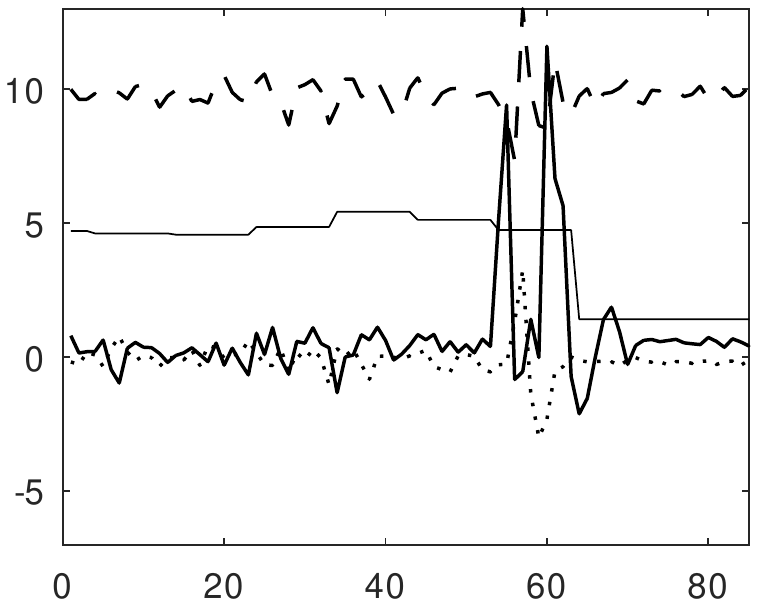}
			\caption{full braking}
			\label{fig:Messreihe_1_Vollbremsung}
		\end{subfigure}

		\vspace{10pt}

		\begin{subfigure}[t]{0.39\textwidth}
			\includegraphics[width=1\textwidth]{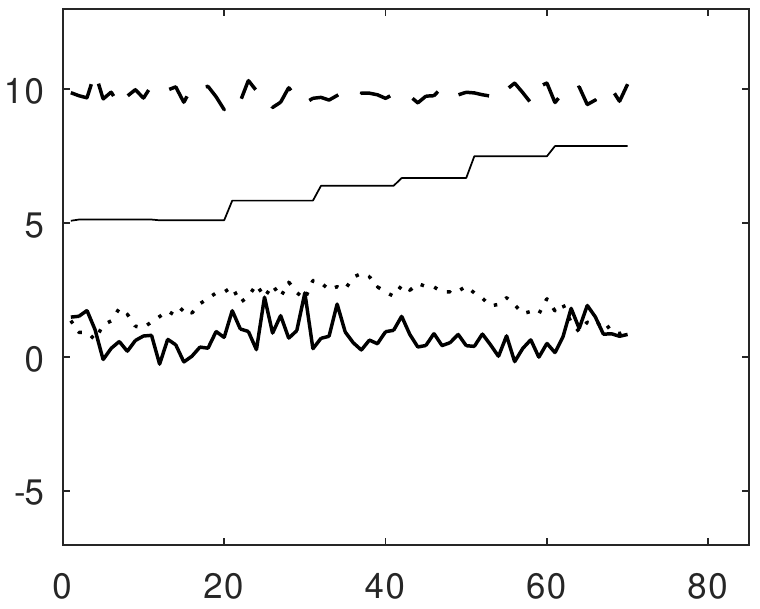}
			\caption{right turn}
			\label{fig:Messreihe_1_Rechtskurve}
		\end{subfigure}
	}
	\begin{subfigure}[t]{0.39\textwidth}
		\includegraphics[width=1\textwidth]{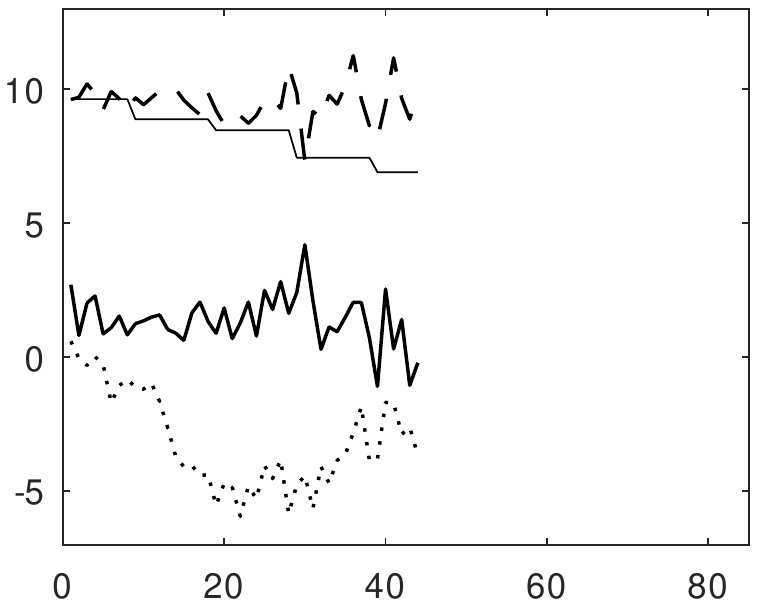}
		\caption{left turn}
		\label{fig:Messreihe_1_Linkskurve}
	\end{subfigure}
\hfill
	\begin{subfigure}[t]{0.19\textwidth}
		\raisebox{19mm}{\includegraphics[width=1\textwidth]{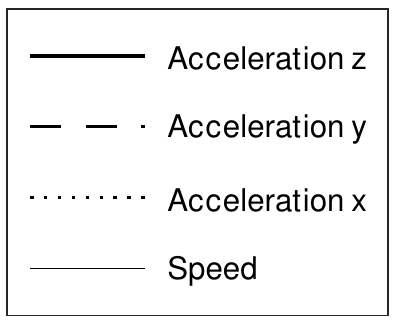}}
	\end{subfigure}

  \caption{Selection of examples of typical maneuver samples (acceleration in
	$\text{m}/\text{s}^2$ and velocity in $\text{m}/\text{s}$). Note that,
	following the lines of \cite{Jae14}, we used only $4$ states (called
	support points) for the conceptor calculation from all measurement series.}
	\label{fig:Messreihen_je_Manoever}
\end{figure}

This test setup was used to collect the required data for seven
different maneuvers. The types of car driving maneuvers and number of measurement series
used for classification are shown in \cref{tab:anzahlManoever}. For the
classification task, only those sections of the measurement series were used
whose features were typical for the car driving maneuver in question. This gives us the
ground truth of the data. Typical measurement series are shown in
\cref{fig:Messreihen_je_Manoever}.
For the classification task, initially only the training data was classified.
For each class~$j$, the conceptors for the classification were calculated with
$8$ measurement series. The number of reservoir neurons $N$ was varied between $2$ and
$60$, and for $N=10$ the number of training examples was varied.
\opt{long}{In these analyses, the available datasets were divided into a
training set and a test set. At the same time, the number of measurement series
used for training varied such that 7 conceptors $C_j^+$ and $C_j^-$ were
calculated for each class~$j$. The smallest training set always consisted of two
samples and was gradually increased until the training set finally consisted of
eight samples.}

\section{Evaluation}

\opt{long}{In this section, we present the results of our isolated word and
speaker recognition task as well as the evaluation of the car driving
maneuvers.}

\subsection{Speech Recognition}

In both speech recognition tasks, we used 50 random ESNs to compare our results.
Task~A (word recognition) was to classify eight different German words. For each
class of words, a positive and a negative conceptor is created (cf.
\cref{evidence}). With them, we calculate the combined evidence (cf. \cref{combined}) to classify test
samples. We use the combined evidence classification, because its accuracy is higher
in comparison to the positive or negative evidence classification alone.
In \cref{taberrorrates}, the error rates of Task~A with and
without conceptors as well as the error rate of a random guess are compared.
With conceptors, our speech recognition system works nearly two times better than an ESN without conceptors. In addition, our method (with conceptors) is about 2.4 times better than a random guess, even if the samples of each class are not that similar.

The goal in Task~B (speaker recognition) was to classify nine
speakers. Without conceptors, we have a mean of 161 false classifications in 50
trials, which equals a mean error rate of 0.544. This misclassification rate
is 3.5 times higher than the mean error rate of the same task with the use of
conceptors.
Our error rate of 0.155 with conceptors in Task~B is much smaller than in
Task~A. The reason probably is that, in Task~A, female and male speakers have
different fundamental voice frequencies pronouncing the same isolated word.
Therefore an analysis that is more invariant to voice frequencies
would be desirable.

\begin{table}\small
  \parbox[t]{.55\linewidth}{\centering
	\caption{Error rates of our two tasks: A.~word recognition and B.~speaker recognition.}
	\label{taberrorrates}\vspace{3mm}
	\begin{tabular}{l|c|c}
		error rates& Task~A & Task~B \\
		\hline
		with conceptors   & 0.366 & 0.155\\
		\hline
		without conceptors  & 0.711 & 0.544\\
		\hline
		random guess  & 0.875 & 0.889
	\end{tabular}}
\hfill
  \parbox[t]{.35\linewidth}{\centering
	\caption{Program runtimes for Task~A and Task~B.}
	\label{timetable}\vspace{3mm}
	\begin{tabular}{l|c|c}
	runtime [s] & Task~A & Task~B \\
	\hline
	training & 0.1704 & 0.1396 \\
	\hline
	test & 0.0057 & 0.0066
	\end{tabular}}
\end{table}

To compare our results with similar speaker recognition tasks using conceptors,
we investigate the speaker classification task in \cite{Jae14} where it is tried
to recognize nine male speakers pronouncing the Japanese di-vowel /ae/. Our dataset
is more diverse in contrast to this dataset, because we use samples of
female and male speakers of different ages. We assume that is the reason why the
mean error rate (combined evidence) 0.062 with the dataset of \cite{Jae14} is
better than the rate of 0.155 for our dataset. We have to be careful directly
comparing the error rates, however, because \cite{Jae14} uses linear predictive
coding (LPC) cepstrum coefficients for data preprocessing, whereas we use MFCCs.
Note that it is not our goal to improve the state-of-the-art word error rate. We
aim to show that we already get good results with a small dataset within very
short training time using ESNs and conceptors. \opt{long}{If the reader is
interested in an accurate automatic speech recognition software for German words
take a look at \cite{MK18}, where 412 training hours for 510 speakers have been
spent.}

In \cref{timetable}, it is shown that the two experiments deliver fast in training
and testing. Our Matlab implementation needs less than $1$\,s for the training
from over 500 samples. This is due to the fact that we used ESNs where the
majority of the weights in the network need not be trained. In comparison to
other neural networks, like convolutional neural networks (CNNs)\opt{long}{ (cf.
\cite{GBC16})} which need thousands of samples and a long training time to get
good results, our study yields proper results already with limited training.
Note that the reported accuracy of CNNs with error rates less than 1\% eventually is
higher. Since the error rate should be low in most speech recognition tasks, a
direct comparison of the proportion of misclassifications with the results of
the presented tasks would be necessary (with consideration of the dataset and
used classification method). Nonetheless, we find using conceptors in both tasks
reduces our error rate many times. This makes conceptors a very powerful means
to improve speech recognition tasks, which should be tested and verified further,
of course.

\subsection{Car Driving Maneuvers}\label{otto}

In this section, we describe our findings with which reservoir size and which
amount of training data classification of car driving maneuvers is possible.
\cref{tab:Klassifikation_der_Trainingsmenge_Neuronen_variabel} shows the results
of classifications with different reservoir sizes. For each class of car
driving maneuvers, $h^+$ and $h^-$ are the positive and negative evidence
vectors used for classification \cite[p.\,77]{Jae14} (cf. \cref{evidence}).
One aim was to find out whether the reservoir size has a significant influence on the classification.
The smallest reservoir we used had 2 neurons, the largest 60 neurons. For each reservoir size, 100 random reservoirs were generated.
The conceptors of positive and negative evidence (cf. \cref{evidence}) were calculated for each driving maneuver and each reservoir.
For each driving maneuver and reservoir, eight measurement series have been used to calculate the conceptors.

As it can be seen in
\cref{tab:Klassifikation_der_Trainingsmenge_Neuronen_variabel}, our
experiments allow classification with very small reservoirs. In the first five
classes, two neurons are sufficient for an almost 100\% correct classification.
Only for the driving maneuvers \enquote{straight ahead} and \enquote{full braking}, the
correctness is only 36\% and 31\% respectively. But the classification accuracy
could be improved with larger reservoirs.
With the exception of \enquote{full braking}, every maneuver is classified
correctly with at least 99\%. In the case of emergency braking, we assume that
significant features of the class are lost due to the previous data processing.
Further investigations with alternative support points (cf. \cref{fig:Messreihen_je_Manoever}) optimized for the class
of emergency braking finally allow significantly better classifications.

\begin{table}\small
\caption{Classification of training data. The table shows the accuracy of the
computed conceptors for positive and negative evidence in \%.}
 \label{tab:Klassifikation_der_Trainingsmenge_Neuronen_variabel}
 \vspace{3mm}
\centering\begin{tabular}{c|c|ccccccccc}
\cline{3-11}
\multicolumn{2}{l|}{}             & \multicolumn{9}{c}{ \textbf{size of reservoir $N$} }                                                                                 \\
\hline
 \textbf{Maneuver}            &    & \textbf{2}  & \textbf{4}  & \textbf{6}  & \textbf{8}  & \textbf{10}  & \textbf{20}  & \textbf{30}  & \textbf{40}  & \textbf{60}   \\
\hline
\multirow{2}{*}{start up}    & $h^+$  & 99.9        & 100         & 100         & 99.9        & 100          & 100          & 99.9         & 100          & 100           \\
\cline{2-2}
                             & $h^-$ & 2.9         & 11.1        & 38.4        & 50.1        & 51.0         & 60.1         & 69.5         & 79.6         & 82.0          \\
\hline
\multirow{2}{*}{slow down}   &  $h^+$  & 99.9        & 99.8        & 99.3        & 98.8        & 99.3         & 99.6         & 99.8         & 99.9         & 100           \\
\cline{2-2}
                             &  $h^-$  & 1.0         & 10.8        & 38.6        & 39.6        & 41.3         & 39.1         & 41.5         & 44.1         & 49.4          \\
\hline
\multirow{2}{*}{right turn} &  $h^+$  & 99.8        & 99.9        & 99.3        & 98.9        & 98.0         & 98.4         & 96.5         & 96.1         & 95.5          \\
\cline{2-2}
                             &  $h^-$  & 78.4        & 92.6        & 97.9        & 98.0        & 96.4         & 100          & 100          & 100          & 100           \\
\hline
\multirow{2}{*}{left turn}  &  $h^+$  & 98.5        & 98.1        & 97.5        & 97.8        & 96.8         & 99.3         & 98.9         & 99.3         & 99.8          \\
\cline{2-2}
                             &  $h^-$  & 36.1        & 36.9        & 79.4        & 92.6        & 93.9         & 98.0         & 100          & 100          & 100           \\
\hline
\opt{long}{\multirow{2}{*}{stop}  &  $h^+$  & 100         & 100         & 100         & 100         & 100          & 100          & 100          & 100          & 100           \\
\cline{2-2}
                             &  $h^-$  & 51.6        & 29.0        & 79.0        & 94.0        & 94.0         & 98.0         & 100          & 100          & 100           \\
\hline}
\multirow{2}{*}{straight ahead}   &  $h^+$  & 36.1        & 62.4        & 89.3        & 93.0        & 99.4         & 100          & 100          & 100          & 100           \\
\cline{2-2}
                             &  $h^-$  & 0.0         & 0.0         & 0.0         & 0.0         & 0.0          & 0.0          & 0.0          & 0.0          & 0.0           \\
\hline
\multirow{2}{*}{full braking}     &  $h^+$  & 31.0        & 34.3        & 40.4        & 45.3        & 47.1         & 54.4         & 59.5         & 60.8         & 64.3          \\
\cline{2-2}
                             &  $h^-$  & 0.0         & 0.0         & 0.0         & 0.0         & 0.0          & 0.0          & 0.0          & 0.0          & 0.0           \\
\hline
\end{tabular}
\end{table}

\cref{poly} (left part) shows the results for classifications with different amounts of training data.
For each driving maneuver, a minimum of two measurements series was used to calculate the conceptors.
The number of measurement series for the training amount was then gradually increased to eight measurement series.
100 random reservoirs were generated and corresponding conceptors were calculated for each class.
This time both the training data and the test data were classified.
Since only a small number of measurement series was available, measurement
series that were not used for training were automatically assigned to the test
set.

Classifications with an accuracy of at least 70.7\% and up to 100\% could be
achieved even with a very small amount of training, consisting of two
measurement series. However, again the driving maneuvers \enquote{straight
ahead} and \enquote{full braking} are an exception and are poorly classified.
Nevertheless, \cref{poly} (left part) also shows that with increasing training
size the accuracy of the classification improves significantly in many cases.

In summary, the accuracy of our car driving maneuver classification is rather
good. Using random guessing as baseline, the accuracy of classifying seven
different car driving maneuvers would reduce to $1/7=14.3$\% which is far less than
our values. Furthermore, in other experiments, often only the simpler task of
binary classification is considered, e.g., only with respect to driver
inattendance \cite{IKL11,TMW04}. In our experiments, we also investigated
open-set testing, i.e., considering also the case that the car driving behavior
does not correspond to one of the given maneuvers. For this, the
thresholds for the components of the evidence vectors have to be adjusted
correctly (see also \cite[Sect.~5.2]{Ott20}).

\begin{table}
\caption{Classification with a varying amount of training data (left part) and
Comparison of the modifications (right part). Both with a constant reservoir
size of $N=10$ neurons. For each driving maneuver, the number of test series $m$
of the test quantity is listed in the line \enquote{test}. The number of
training series $n$ is listed at the top of the table.}
\label{poly}\vspace{4mm}
	\small
	\centering
\begin{tabular}{c|c|c||ccccccc||cc|cc|cc}
\cline{4-16}
\multicolumn{2}{c}{ }                 &      &  \multicolumn{7}{c||}{ \textbf{classification}}  & \multicolumn{6}{c}{ \textbf{modifications }}\\
\cline{4-16}
\multicolumn{2}{c}{}                 &         & \multicolumn{7}{l||}{ \textbf{}}	& \multicolumn{2}{c|}{\textbf{linear } } & \multicolumn{2}{c|}{\textbf{polynomial} } & \multicolumn{2}{l}{\begin{tabular}[c]{@{}l@{}} \textbf{linear and}\\ \textbf{polynomial}\end{tabular}} \\
\hline
\multicolumn{2}{l}{\textbf{} } & $n$  &     \textbf{2 } & \textbf{3 } & \textbf{4}  & \textbf{5 } & \textbf{6 } & \textbf{7}  & \textbf{8 }  & \textbf{5}     & \textbf{8}            & \textbf{5}    & \textbf{8}           & \textbf{5}     & \textbf{8}                                                                \\
\hline
\multirow{5}{*}{\shortstack{start\\ up}}       & \multirow{2}{*}{training} & $h^+$ & 100  & 100  & 100  & 100  & 100  & 99.7 & 99.9 & 1.0            & 0.5            & 0              & 0              & 0.8            & 0.9                                                               \\
\cline{3-3}
                                &                           & $h^-$    & 93.0 & 74.0 & 63.5 & 47.6 & 48.5 & 53.4 & 52.1 & \textbf{-4.0}  & \textbf{-6.8}  & 0              & \textbf{-0.4}  & \textbf{-4.6}  & \textbf{-5.8}                                                   		 \\
\cline{2-3}
                                & \multirow{3}{*}{test}     & m       & 9    & 8    & 7    & 6    & 5    & 4    & 3& 6              & 3              & 6              & 3              & 6              & 3                                                                				\\
\cline{3-3}
                                &                           & $h^+$   & 83.7 & 97.3 & 98.9 & 100  & 100  & 100  & 100 & 0              & 0              & 0              & 0              & 0              & 0                                                                 \\
\cline{3-3}
                                &                           & $h^-$     & 37.7 & 54.6 & 50.6 & 49.3 & 69.8 & 72.5 & 88.0 & \textbf{-6.0}  & \textbf{-1.3}  & \textbf{-0.3}  & \textbf{-1.0}  & \textbf{-7.7}  & 1.7                                                             \\
\hline
\multirow{5}{*}{\shortstack{slow\\ down}} & \multirow{2}{*}{training} & $h^+$  & 100  & 100  & 100  & 100  & 100  & 98.4 & 99.1  & 0              & 2.4            & 0              & \textbf{-0.6}  & 0              & 1.5                                                              \\
\cline{3-3}
                                &                           & $h^-$   & 49.5 & 56.7 & 63.0 & 58.8 & 45.3 & 44.9 & 46.8  & 23.2           & 29.0           & 3.4            & 4.9            & 21.0           & 23.4                                                            \\
\cline{2-3}
                                & \multirow{3}{*}{test}     & m     & 6    & 5    & 4    & 3    & 2    & 1    & 0& 3              & 0              & 3              & 0              & 3              & 0                                                                 \\
\cline{3-3}
                                &                           &  $h^+$      & 88.5 & 91.0 & 88.0 & 88.7 & 85.5 & 100  & - & 14.0           & -              & \textbf{-1.7}  & -              & 5.7            & -                                                              \\
\cline{3-3}
                                &                           &  $h^-$    & 37.7 & 29.4 & 11.0 & 6.7  & 7.0  & 27.0 & -& 1.3            & -              & 2.3            & -              & 1.7            & -                                                               \\
\hline
\multirow{5}{*}{\shortstack{right\\ turn}}    & \multirow{2}{*}{training} & $h^+$   & 100  & 98.7 & 97.0 & 97.8 & 99.5 & 96.4 & 98.5 & 2.2            & 0.3            & \textbf{-0.4}  & \textbf{-0.5}  & 1.8            & 1.6                                                               \\
\cline{3-3}
                                &                           &  $h^-$    & 98.0 & 99.0 & 96.0 & 100  & 99.3 & 98.0 & 96.8& 1.2            & \textbf{-1.5}  & 1.6            & \textbf{-0.6}  & 3.0            & 0.1                                                              \\
\cline{2-3}
                                & \multirow{3}{*}{test}     & m    & 15   & 14   & 13   & 12   & 11   & 10   & 9  & 12             & 9              & 12             & 9              & 12             & 9                                                                \\
\cline{3-3}
                                &                           &  $h^+$    & 98.5 & 97.4 & 95.6 & 97.7 & 99.1 & 97.5 & 99.2 & 1.8            & \textbf{-0.3}  & 0              & 0.4            & 1.3            & 0.1                                                             \\
\cline{3-3}
                                &                           &  $h^-$     & 94.1 & 96.6 & 95.2 & 97.5 & 97.1 & 97.0 & 96.0& 1.5            & \textbf{-3.0}  & 1.7            & \textbf{-1.0}  & 1.4            & \textbf{-1.0}                                                 \\
\hline
\multirow{5}{*}{\shortstack{left\\ turn}}     & \multirow{2}{*}{training} & $h^+$   & 100  & 99.7 & 100  & 100  & 99.8 & 95.6 & 96.4 & 0              & \textbf{-0.9}  & 0              & \textbf{-0.6}  & 0              & 0.4                                                               \\
\cline{3-3}
                                &                           &  $h^-$    & 94.0 & 95.7 & 95.0 & 96.0 & 95.0 & 93.0 & 93.9& 2.8            & \textbf{-0.6}  & 0.2            & \textbf{-0.6}  & 3.0            & \textbf{-0.4}                                                   \\
\cline{2-3}
                                & \multirow{3}{*}{test}     & m      & 13   & 12   & 11   & 10   & 9    & 8    & 7 & 10             & 7              & 10             & 7              & 10             & 7                                                                \\
\cline{3-3}
                                &                           &  $h^+$    & 70.7 & 85.8 & 87.6 & 83.0 & 91.6 & 98.4 & 99.4 & 1.4            & 0              & 0              & 0              & 0.9            & \textbf{-0.1}                                                   \\
\cline{3-3}
                                &                           &  $h^-$    & 93.0 & 94.6 & 95.0 & 95.2 & 95.0 & 93.0 & 94.0  & 1.9            & \textbf{-1.0}  & 1.1            & \textbf{-1.0}  & 2.2            & \textbf{-1.0}                                                   \\
\hline
\opt{long}{\multirow{5}{*}{stop}   & \multirow{2}{*}{training} & $h^+$     & 100  & 100  & 100  & 100  & 100  & 100  & 100 & 0              & 0              & 0              & 0              & 0              & 0                                                               \\
\cline{3-3}
                              &                           &  $h^-$  & 94.0 & 96.0 & 95.0 & 95.0 & 95.0 & 94.0 & 95.0 & 6.0            & \textbf{-1.0}  & 0              & \textbf{-1.0}  & 4.0            & 0                                                                 \\
\cline{2-3}
                              & \multirow{3}{*}{test}     & m      & 8    & 7    & 6    & 5    & 4    & 3    & 2 & 5              & 2              & 5              & 2              & 5              & 3                                                               \\
\cline{3-3}
                              &                           &  $h^+$      & 100  & 100  & 100  & 100  & 100  & 100  & 100& 0              & 0              & 0              & 0              & 0              & 0                                                               \\
\cline{3-3}
                              &                           &  $h^-$  & 94.0 & 96.0 & 95.0 & 95.0 & 95.0 & 94.0 & 95.0  & 6.0            & \textbf{-1.0}  & 0              & \textbf{-1.0}  & 4.0            & 0                                                               \\
\hline}
\multirow{5}{*}{\shortstack{straight\\ ahead}}    & \multirow{2}{*}{training} & $h^+$  & 93.5 & 95.7 & 99.5 & 95.0 & 97.7 & 96.6 & 95.6 & 0.6            & 2.9            & \textbf{-2.6}  & \textbf{-1.9}  & \textbf{-2.0}  & 3.8                                                             \\
\cline{3-3}
                              &                           & $h^-$  & 0    & 0    & 0    & 0    & 0    & 0    & 0 & 0              & 0              & 0              & 0              & 0              & 0                                                                 \\
\cline{2-3}
                              & \multirow{3}{*}{test}     & m     & 7    & 6    & 5    & 4    & 3    & 2    & 1  & 4              & 1              & 4              & 1              & 4              & 1                                                               \\
\cline{3-3}
                              &                           &  $h^+$    & 45.4 & 74.3 & 89.6 & 82.3 & 80.0 & 74.5 & 48.0 & 2.0            & 7.0            & \textbf{-4.3}  & \textbf{-5.0}  & \textbf{-2.3}  & 8.0                                                              \\
\cline{3-3}
                              &                           & $h^-$   & 0    & 0    & 0    & 0    & 0    & 0    & 0 & 0              & 0              & 0              & 0              & 0              & 0                                                                 \\
\hline
\multirow{5}{*}{\shortstack{full\\ braking}} & \multirow{2}{*}{training} & $h^+$  & 87.0 & 63.0 & 30.8 & 34.2 & 32.8 & 40.0 & 46.3 & \textbf{-1.4}  & 1.5            & 0.8            & \textbf{-2.1}  & \textbf{-1.6}  & 0.6                                                             \\
\cline{3-3}
                              &                           &  $h^-$   & 22.0 & 2.3  & 0    & 0    & 0    & 0    & 0  & \textbf{-0.2}  & 0              & 0              & 0              & 0              & 0                                                             \\
\cline{2-3}
                              & \multirow{3}{*}{test}     & m    & 6    & 5    & 4    & 3    & 2    & 1    & 0   & 3              & 0              & 3              & 0              & 3              & 0                                                               \\
\cline{3-3}
                              &                           &  $h^+$    & 11.0 & 35.6 & 20.3 & 36.7 & 78.5 & 69.0 & - & 0.7            & -              & 3.3            & -              & 4.7            & -                                                               \\
\cline{3-3}
                              &                           &  $h^-$  & 0    & 0    & 0    & 0    & 0    & 0    & - & 0              & -              & 0              & -              & 0              & -                                                               \\
\hline
\end{tabular}
\end{table}

\opt{long}{\vspace*{-1pt}}
\subsection{Identifying Essential Factors}\label{essen}

Let us finally analyze essential factors of car driving maneuver detection,
namely the influence of 1.~using a linear activation function and
2.~omitting polynomial interpolation, which are part of the method in
\cite{Jae14}, on the classification accuracy:
\vspace{-0.17cm}
\begin{description}
  \item[linear vs. nonlinear activation function:] Motivated by linear
	neural networks \cite{SL+18} where all neurons are linearly activated,
	experiments with linear activation were carried out. Only the activation
	function of the existing classifier was changed for these analyses.
	Instead of $\tanh$, a linear activation function is applied.
  \item[with vs. without polynomial interpolation:] We have analyzed the
	influence of polynomial interpolation on the classifications of driving
	maneuvers. For this purpose, polynomial interpolation was not performed
	after data normalization. Here again $\tanh$ was used as the
	activation function in these analyses.
\end{description}

In total, $100$ test runs (each with a randomly generated reservoir) with $n=5$
and $n=8$ training data sequences were performed. In \cref{poly} (right part),
$h^+$ and $h^-$ again are the positive and negative evidence
vectors used for classification (cf. \cref{otto}). We can see how much
the classification with linear activation is better or worse than the
classification with $\tanh$ as activation function. In the column \enquote{polynomial} it is shown
how much the classification is better or worse if polynomial interpolation
is not performed. The last column shows the results if both
linear activation and no polynomial interpolation is performed. Results for the
classification of training data and test data are presented separately for each car
driving maneuver. The difference between the classification of the original
implementation (with polynomial interpolation and the modified implementation)
is given in percent. The calculation rule is: $\Delta_\text{quality} =
\text{quality}_\text{original} - \text{quality}_\text{modification}$. If
$\Delta_\text{quality}$ is negative, the classification of the original
implementation is worse. These values are highlighted in \cref{poly} (right part).

Looking at the classification of the training measurement series, the
classification with linear activation is 2.9\% worse in the worst case (\enquote{straight
ahead}, $h^+$, $n=8$). For emergency braking ($h^+$, $n=5$) even better by 1.4\%.
For $h^-$ there are also opposing values. For \enquote{start up}, the linear
activation achieves a better classification (4.0\% or 6.8\% better), whereas for
\enquote{slow down} it achieves a significantly worse classification (23.2\% or
29.0\% worse). Finally, when classifying the training set with a linear
activation for $h^+$ almost always the same quality is achieved as by the
original implementation with~$\tanh$.

Polynomial interpolation seems to have less impact on the classification.
Although there are again classifications that achieve a worse quality without
polynomial interpolation. Overall the quality of the modification is up to 5\%
better. The greatest (positive) deterioration compared to the original implementation is
4.9\% for the \enquote{slow down} class (training measurement series, $h^-$, $n=8$).
Often even the same quality is achieved ($\Delta_\text{quality} = 0$).
For further details the reader is referred to \cite{Ott20}.

In conclusion, it should be noted that no uniform result could be achieved, thus
a generally positive or negative effect can not be shown. The original and
the modified implementations achieve similar quality. But from this we may
conclude that the commonly used hyperbolic tangent function and polynomial
interpolation is not really mandatory to solve classification tasks. Hence the
procedure in \cite{Jae14} can be simplified significantly by using plain linear
activation and omitting polynomial interpolation.

\section{Conclusions}

We have shown that fast learning with ESNs and conceptors is possible for
diverse applications, in particular, speech recognition and detecting car driving
maneuvers. With already hardly more than one hundred samples per class accurate
classification is possible, in contrast to other neural networks like CNNs that
require thousands of samples for each class. In addition, training conceptor
networks for classification lasts only a few seconds. Nevertheless only with
application-specific methods high accuracy is reachable, e.g., using MFCCs for
speech recognition. Furthermore, existing procedures can be simplified, e.g., employing simple linear activation and
omitting polynomial interpolation without loosing much accuracy, as done for car
driving maneuver detection. Further work will consider even more applications
and simplifications of RNNs (cf. \cite{SL+18}) and focus more on
open-set testing, which seems to be important also for speech recognition
\cite{CPT06}.
While we use a common one-vs-all classification scheme with as many output units
as there are classes in a dataset, a novel one-vs-one classification scheme that
trains each output unit to distinguish between a specific pair of classes had
recently been developed \cite{PO+20}. A comparison to our classification method
could be of further interest.

\bibliographystyle{splncs04}
\bibliography{conceclass}

\end{document}